\documentclass[10pt,twocolumn,letterpaper]{article}

\usepackage{cvpr}              
\usepackage{algorithm}
\usepackage{algpseudocode}
\usepackage{colortbl}
\usepackage{soul}
\usepackage{graphicx} 
\usepackage{subcaption} 
\usepackage{svg}
\usepackage{bibunits} 

\definecolor{cvprblue}{rgb}{0.21,0.49,0.74}
\usepackage[pagebackref,breaklinks,colorlinks,allcolors=cvprblue]{hyperref}

\def\paperID{5975} 
\def\confName{CVPR}
\def\confYear{2025}

\title{``Stones from Other Hills can Polish Jade'':\\
Zero-shot Anomaly Image Synthesis via Cross-domain Anomaly Injection}

\author{
  Siqi Wang\textsuperscript{1*}, Yuanze Hu\textsuperscript{1*}, Xinwang Liu\textsuperscript{1}, 
  Siwei Wang\textsuperscript{2}, Guangpu Wang\textsuperscript{1}, Chuanfu Xu\textsuperscript{1}, 
  Jie Liu\textsuperscript{1}, Ping Chen\textsuperscript{3}\\
  \textsuperscript{1}National University of Defense Technology, Changsha, China\\
  \textsuperscript{2}Academy of Military Science\\
  \textsuperscript{3}North University of China\\
}

\begin{document}

\maketitle
\begin{abstract}
Industrial image anomaly detection (IAD) is a pivotal topic with huge value. Due to anomaly’s nature, real anomalies in a specific modern industrial domain (i.e.\enspace domain-specific anomalies) are usually too rare to collect, which severely hinders IAD. Thus, zero-shot anomaly synthesis (ZSAS), which synthesizes pseudo anomaly images without any domain-specific anomaly, emerges as a vital technique for IAD. However, existing solutions are either unable to synthesize authentic pseudo anomalies, or require cumbersome training. Thus, we focus on ZSAS and propose a brand-new paradigm that can realize both authentic and training-free ZSAS. It is based on a chronically-ignored fact: \textbf{Although domain-specific anomalies are rare, real anomalies from other domains (i.e.\enspace cross-domain anomalies) are actually abundant and directly applicable to ZSAS.} Specifically, our new ZSAS paradigm makes three-fold contributions: First, we propose a novel method named Cross-domain Anomaly Injection (CAI), which directly exploits cross-domain anomalies to enable highly authentic ZSAS in a training-free manner. Second, to supply CAI with sufficient cross-domain anomalies, we build the first Domain-agnostic Anomaly Dataset within our best knowledge, which provides ZSAS with abundant real anomaly patterns. Third, we propose a CAI-guided Diffusion Mechanism, which further breaks the quantity limit of real anomalies and enable unlimited anomaly synthesis. Our head-to-head comparison with existing ZSAS solutions justifies our paradigm’s superior performance for IAD and demonstrates it as an effective and pragmatic ZSAS solution. 
\end{abstract}    
\section{Introduction}
\label{sec:intro} 

As a long-standing topic, industrial image anomaly detection (IAD) \cite{liu2024deep} has constantly been attractive due to its potential to automate the tedious visual inspection process in the industrial pipeline. However, IAD remains challenging despite of significant dedicated effort, which can be attributed to the following reason: Real anomalies in a specific industrial domain (called \textbf{\textit{domain-specific anomalies}} in this paper) are often rare due to the highly standardized feature of modern industrial pipeline \cite{pang2021deep}. Such rarity makes it hard or even impossible to collect domain-specific anomalies as training data, which hinders one from addressing IAD by a mature and frequently-used supervised classification paradigm. 
\begin{figure}[t]
  \centering
   \includegraphics[width=0.95\linewidth]{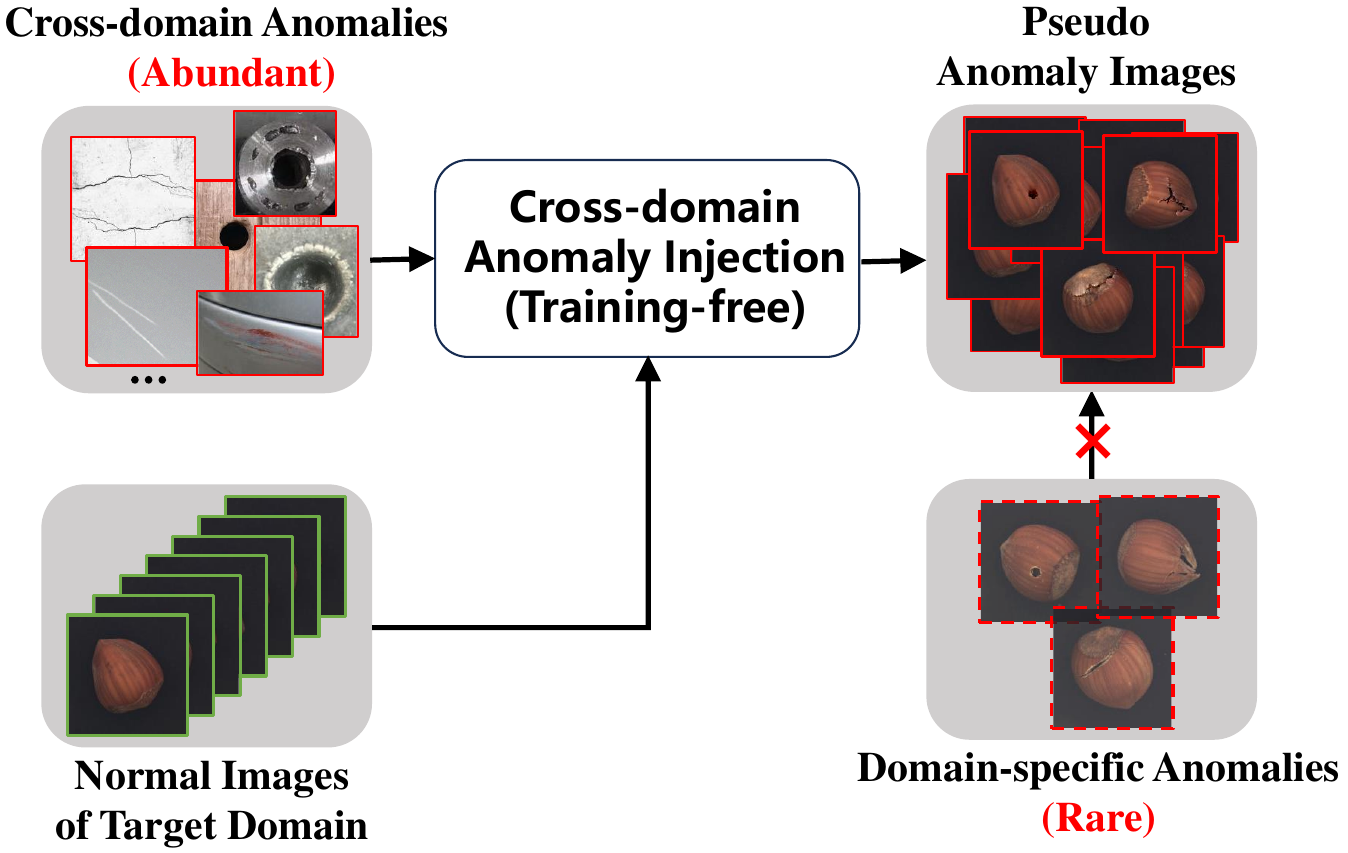} 
   \caption{ZSAS flow of Cross-domain Anomaly Injection.}
   \label{fig:flow_cai}
\end{figure}
As a result, the literature generally studies IAD under a \textit{one-class classification} (OCC) setting that assumes no domain-specific anomaly is available. The OCC setting inspires most IAD solutions to discard all clues from the anomaly class during training. Instead, those solutions focus on building a normality model from purely normal training images, while images with evident deviations are inferred as anomalies.  Although no real anomaly is required anymore, recent research reveals that such OCC-oriented IAD solutions still face vital problems caused by completely ignoring the anomaly class, e.g. inability to learn IAD-related features that are sufficiently discriminative \cite{schluter2022natural}, difficulty in detecting subtle anomalies \cite{hu2024anomalydiffusion}, etc.

To ease above problems, recent years have witnessed a surging interest in few-shot/zero-shot anomaly synthesis (FSAS/ZSAS), which aims to synthesize pseudo anomaly images for a specific industrial domain with few or zero domain-specific anomaly. A fair number of FSAS/ZSAS solutions have been proposed (reviewed in Sec. \ref{sec:Related-Work}), while pseudo anomalies gradually occupy an indispensable role in many recent IAD solutions \cite{zavrtanik2021draem,li2021cutpaste,schluter2022natural,zhang2024realnet,hu2024anomalydiffusion}. Since FSAS still requires accessing a few domain-specific anomalies in advance, FSAS suffers from the uncertainty or even unavailability of domain-specific anomalies. By contrast, ZSAS makes no assumption on the availability of domain-specific anomalies, thus enjoying better applicability to more diverse domains. Hence, we will focus on ZSAS in this paper. To maximize pseudo anomalies' utility and facilitate them to be incorporated into IAD, a natural goal is to realize both \textbf{\textit{authentic and simple ZSAS}}. Unfortunately, we observe that existing ZSAS solutions find it hard to balance authenticity and simplicity. Specifically, existing ZSAS solutions can be categorized into {\textit{model-free} solutions and {\textit{model-based} solutions: Model-free solutions (e.g. CutPaste \cite{li2021cutpaste}, DRAEM \cite{zavrtanik2021draem}) play a dominant role in ZSAS. They sample patterns from given normal images or external datasets that are irrelevant to IAD, and directly inject those patterns into the target normal image to synthesize anomalies. In this way, such solutions enjoy a simple use without training an extra model. However, pseudo anomalies synthesized by model-free solutions typically lack authenticity, because their pseudo anomaly patterns are irrelevant to real anomalies. Besides, their authenticity is further diminished by injection artifacts (e.g. pixel discontinuity). By contrast, recent model-based solutions like RealNet \cite{zhang2024realnet} place more emphasis on authenticity. They promote authenticity by training a deep generative model like Generative Adversarial Networks (GANs) \cite{goodfellow2020generative} or Diffusion Models (DMs) \cite{song2019generative}, while the core idea is to re-direct the model's generative process by some external factors (e.g. carefully-crafted perturbations) to generate pseudo anomalies. Although authenticity is improved, such solutions involve a cumbersome training process that requires heavy calculations and extra hardware. Meanwhile, re-directing the generative process is not straightforward, which often requires delicate model design and tuning. Therefore, model-based solutions lack simplicity due to their training-based nature, and their high cost limits their application to practical IAD.

In this paper, we are naturally motivated to develop a new ZSAS paradigm that can attain authenticity and simplicity at the same time. Unlike previous OCC-oriented ZSAS methods that recklessly ignore ALL real anomalies, we first point out a chronically-ignored fact in IAD: Although domain-specific anomalies are often too rare to be collected in a certain industrial domain, \textbf{\textit{we actually enjoy access to abundant real anomalies from other domains, i.e. cross-domain anomalies.}} In fact, cross-domain anomalies have been accumulated for years, and they can be easily acquired from multiple sources, such as search engines and existing datasets. Then, as the Chinese proverb ``stones from other hills can polish jade'' goes, we argue that \textbf{\textit{a fair amount of cross-domain anomalies actually provide rich and sufficient real anomaly patterns to enable authentic and simple ZSAS}}. Thus, as shown in Fig. \ref{fig:flow_cai}, we propose a brand-new ZSAS paradigm that contributes in three ways:

\begin{itemize}
	\item We propose a novel anomaly synthesis method named Cross-domain Anomaly Injection (CAI), which can exploit abundant cross-domain anomalies to directly synthesize highly authentic synthetic anomalies in a training-free manner. In this way, CAI for the first time achieves authenticity and simplicity simultaneously in ZSAS.
	\item We build the very first domain-agnostic anomaly dataset (DAAD) within our best knowledge to supply CAI with rich real anomaly patterns from abundant cross-domain anomalies. It also benefits future ZSAS/IAD research.
	\item We propose a CAI-guided Diffusion Mechanism, which can utilize CAI's authentic pseudo anomaly images to perform unlimited anomaly synthesis. It enables us to further break the quantity limit of real anomalies in DAAD.
    
\end{itemize}

Comparison with state-of-the-art (SOTA) ZSAS solutions justifies the superiority of our paradigm in IAD, and it can serve as a simple and pragmatic solution in practice.

\section{Related Work}
\label{sec:Related-Work}

\textbf{Industrial Image Anomaly Detection (IAD).} IAD has constantly been an active research topic with vast solutions, and they basically fall into the following categories: (1) \textit{Reconstruction-based methods} usually learn to reconstruct normal image patterns from given normal images \cite{gong2019memorizing} or pseudo anomaly images \cite{zavrtanik2021draem, schluter2022natural} by a deep neural network (DNN), while large reconstruction errors are usually used to signify anomalies.
(2) \textit{Classification-based methods} either introduce pseudo anomalies to transform IAD into a binary classification problem \cite{li2021cutpaste,liu2023simplenet}, or directly address IAD by OCC methods \cite{yi2020patch,reiss2021panda}.
(3) \textit{Distillation-based methods} usually fix a pre-trained DNN as the teacher network, and learn to distill key knowledge of normal images into a student DNN \cite{bergmann2020uninformed,zhang2023destseg,liu2024dual}. Then anomalies are detected by measuring how well the knowledge is transferred. (4) \textit{Distribution-based methods} learn to map normal images to a target distribution \cite{rudolph2021same,rudolph2022fully}, while images that divert from the distribution after mapping are viewed as anomalies.  
(5) \textit{Embedding-based methods} \cite{fang2023fastrecon} typically utilize a pre-trained DNN backbone to extract feature embeddings from given normal images, which are then used as templates to detect distant anomalous embeddings. A detailed review of IAD can be found in \cite{liu2024deep}. In particular, it is noted that anomaly synthesis plays an increasingly important role, which facilitates IAD to learn more discriminative representations \cite{li2021cutpaste}, enhance subtle anomaly detection \cite{zhang2023destseg}, narrow domain gap between pre-training tasks and IAD \cite{zhang2024realnet}, etc.

\textbf{Anomaly Synthesis (AS).}
In IAD, AS can be divided into FSAS and ZSAS. FSAS adopts a relaxed setting that assumes availability of few domain-specific anomalies, and clues from those domain-specific anomalies are used as conditions for modern generative DNNs like GANs \cite{li2022eid,zhang2021defect} or DMs \cite{hu2024anomalydiffusion,jiang2024cagen}. As we mentioned in Sec. \ref{sec:intro}, pre-conditions of FSAS may not always be satisfied, while it suffers from uncertain number and type of accessible domain-specific anomalies. By contrast, ZSAS is more challenging, since it does not probe any domain-specific anomalies in prior. The absence of domain-specific anomalies encourages most solutions to perform ZSAS in a model-free fashion. Specifically, model-free ZSAS samples existing image patterns from given normal images \cite{li2021cutpaste,schluter2022natural,zou2022spot} or external IAD-irrelevant datasets \cite{zavrtanik2021draem} as pseudo anomalies, which are directly injected into normal images from target industrial domain by multiple means, e.g. pasting \cite{li2021cutpaste}, smooth blending \cite{zou2022spot} or Poisson Editing (PE) \cite{10.1145/1201775.882269}. To remedy the authenticity problem of model-free ZSAS, model-based ZSAS methods are also proposed. They focus on re-directing the generative process to generate images that divert from normality as pseudo anomalies. The re-direction can be realized by adding carefully-designed perturbations \cite{zhang2024realnet} or introducing clues from limited real anomalies as conditions \cite{rippel2020gan}. In particular, we notice that \cite{rippel2020gan} also leverage real anomalies from other domains for anomaly synthesis, but our work significantly differs from \cite{rippel2020gan}: Our work points out the abundant nature of cross-domain anomalies, which then enables us to conduct training-free ZSAS by a direct cross-domain anomaly injection process. By contrast, \cite{rippel2020gan} ignores the abundance of cross-domain anomalies. Instead, it trains a GAN with few real anomalies for indirect anomaly synthesis, and it is only applied to the fabric domain.

\begin{figure}[t]
  \centering
   \includegraphics[width=0.925\linewidth]{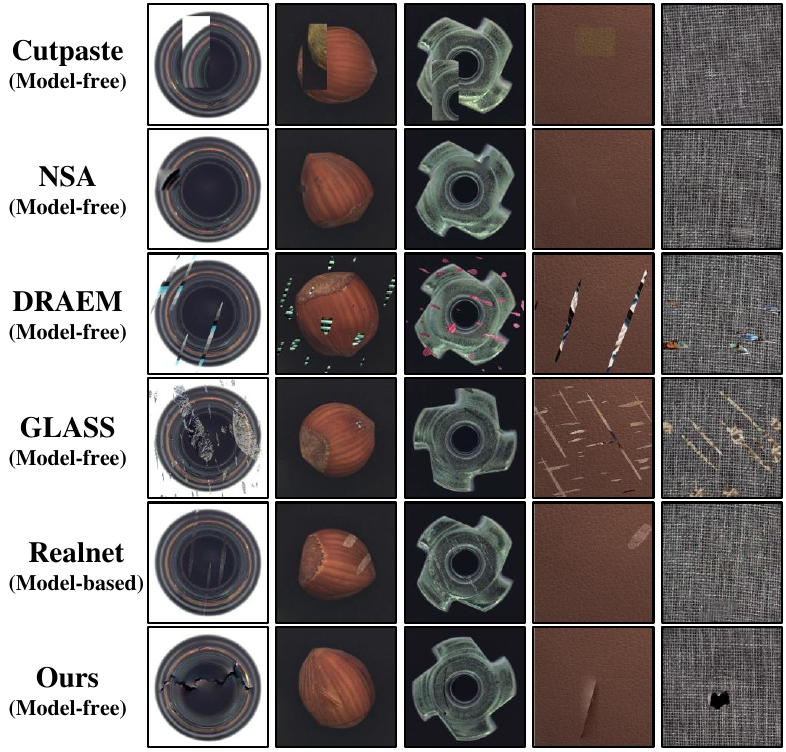} 
   \caption{Comparison of pseudo anomaly images synthesized by SOTA ZSAS solutions and the proposed CAI. }
   \label{fig:zsas_compare}
\end{figure}

\section{The Proposed ZSAS Paradigm}

To tackle the deficiency of existing ZSAS solutions, we develop a new ZSAS paradigm that can strike a good balance between simplicity and authenticity. In this section, we detail three key components of our paradigm: The core CAI method for ZSAS, a new domain-agnostic anomaly dataset as the data foundation of CAI, and a diffusion-based mechanism that enables further extension of pseudo anomalies.

\subsection{Cross-domain Anomaly Injection (CAI)}

Our ZSAS method is motivated by two vital observations: (1) 
Cross-domain anomalies are actually abundant in practice, so we can readily access considerable real anomaly patterns. (2) Despite of the difference in industrial domains, anomaly patterns in one anomaly class are often similar (e.g. cracks), which is also observed by prior works \cite{wang2023defect,zhouanomalyclip}. Based on two observations above, cross-domain anomalies provide an excellent source of abundant real anomaly patterns. As those patterns are real in themselves, they can be readily transferred across domains to guarantee the authenticity of pseudo anomalies, and provide valuable clues of anomalies to assist IAD. Meanwhile, we argue that cross-domain anomalies' abundance enables massive anomaly synthesis by direct anomaly injection, which makes it no longer necessary to  train a generative model. In this way, we propose a novel method named Cross-domain Anomaly Injection (CAI), which achieves both simple and authentic ZSAS (see Fig. \ref{fig:zsas_compare}). Before introducing CAI, we first formulate the general ZSAS procedure of CAI below:

Given a set of cross-domain anomalies, we randomly sample a cross-domain anomaly image $I_i^{(ca)}\in \mathbb{R}^{H_a\times W_a\times 3}$ as source image, and its binary mask $y_i\in \{0, 1\}^{H_a\times W_a}$ that indicates the area of its anomaly pattern $A_i=I_i^{(ca)}(y_i)$, where $I_i^{(ca)}(y_i)$ refers to the image pattern indexed by $y_i$'s non-zero elements. Then, we fetch a target image $I^{(n)}\in \mathbb{R}^{H_n\times W_n\times 3}$ from the normal training set of target domain and resize it to the target size for processing. Finally, CAI intends to inject $A_i$ into $I^{(n)}$ to synthesize a pseudo anomaly image $I^{(sa)}$. Concretely, CAI consists of four major stages: Source-target Matching, Multi-scale Anomaly Synthesis, Injection Location Selection and PE-based Anomaly Injection. The procedure of CAI is shown in Fig. \ref{fig:CAI} and Algo. \ref{alg:cai}, while we elaborate each stage below:

\begin{figure}[t]
  \centering
   \includegraphics[width=0.88\linewidth]{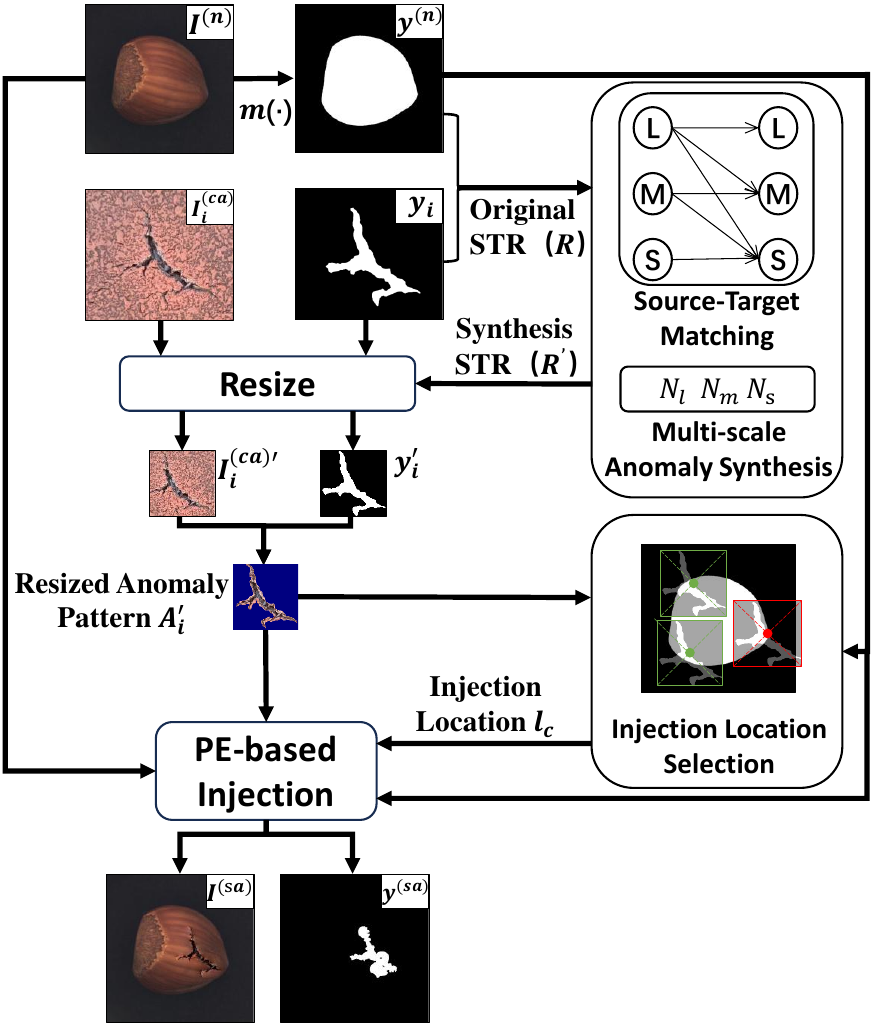} 
   \caption{Detailed procedure of CAI for ZSAS.}
   \label{fig:CAI}
\end{figure}

\textbf{Source-target Matching (STM).} As the source image $I_i^{(ca)}$ and target image $I^{(n)}$ are from different domains, real anomaly pattern $A_i$ in  $I_i^{(ca)}$ often requires resizing before injection to fit $I^{(n)}$. In this process, we need to avoid improper resizing, which makes injected anomaly patterns too trivial or too vague. To this end,  we propose Source-target Matching. Specifically, we first extract a binary mask to indicate the foreground in $I^{(n)}$: $y^{(n)}=m(I^{(n)})$, where the function $m(\cdot)$ either returns the mask of the object by image matting \cite{gatis_rembg_2022}, or return an all-one mask for images without any object (e.g. texture). Afterwards, we calculate the {scale} of the anomaly pattern in $I_i^{(ca)}$ and the foreground in $I^{(n)}$, and define the original \textit{Source-Target Ratio} (STR), $R$:

\begin{equation}
\small
\label{eq:str}
\begin{aligned}
       s_a, s_f = max(minbb(y_i)), max(minbb(y^{(n)})), R =\frac{s_a}{s_f} \\
\end{aligned}
\end{equation}
where $minbb(\cdot)$ returns the height and width of the minimum bounding box that can enclose the input mask's non-zero pixels. $R$ aims to measure the original scale of $I_i^{(ca)}$'s anomaly pattern for injection. With $R$ as an indicator, we divide the anomaly pattern of source image into four scale types: Trivial ($R<0.1$), small ($R\in[0.1, 0.3)$), medium ($R\in[0.3, 0.7)$) and large ($R\geq 0.7$). Similarly, we can divide pseudo anomalies into three scale types by the \textit{Synthesis STR}, $R'$: Suppose that a pseudo anomaly image contains an injected anomaly pattern $A'$ with a scale $s'_a=max(minbb(A'))$, so $R'$ is defined by $R'=s'_a/s_f$. Then, we can classify pseudo anomalies into three scale types:
Small ($R'\in [0.1, 0.3)$), medium ($R'\in [0.3, 0.7)$) and large ($R'\in [0.7, 1]$). To exploit the original $A_i$ to synthesize a pseudo anomaly with Synthesis STR $R'$, we can proportionally resize $(I_i^{(ca)}, y_i)$ to $(I_i^{(ca)}{}', y'_i)$ by the ratio $r=R'/R$. Since an improper $R'$ for resizing will make the resized anomaly pattern $A'_i=I_i^{(ca)}{}'(y'_i)$ too trivial or too vague, STM imposes the following constraints to choose $R'$: (1) Trivial anomaly patterns are discarded. (2) Anomaly patterns cannot be resized to synthesize pseudo anomalies of a larger scale. For example, a medium pattern can be used to synthesize medium or small pseudo anomalies, but it cannot be used to produce large pseudo anomalies. In this way, STM ensures the visibility and clarity of injected anomaly patterns, which are vital to authentic pseudo anomalies.

\begin{algorithm}[t]
\small
    \caption{Cross-domain Anomaly Injection.}
    \label{alg:cai}
    \textbf{Input}: $\mathcal{I}^{(ca)}$, $I^{(n)}$, $N_l, N_m, N_s$\\
    \textbf{Output}: A set of pseudo anomaly images $\mathcal{I}^{(sa)}$.
    \begin{algorithmic}[1] 
        \State Initialize $\mathcal{I}^{(sa)}=\{\}$, counters $C_l=0, C_m=0, C_s=0$. 
        \State Get $I^{(n)}$'s foreground mask $y^{(n)}=m(I^{(n)})$. 
        \While{$C_l<N_l$ or $C_m<N_m$ or $C_s<N_s$} 
        \State Randomly sample $(I_i^{(ca)}, y_i)\in\mathcal{I}^{(ca)}$.
        \State Calculate the original STR $R$ by Eq. (\ref{eq:str}).
        \If{$R<0.1$} \textbf{continue}.
        \ElsIf{$R\geq 0.7$}
        \If{$N_l<C_l$} 
        \State Select a random $R'\in[0.7,1]$, $C_l=C_l+1$.
        \ElsIf{$N_m<C_m$} 
        \State Select a random $R'\in[0.3,0.7)$, $C_m=C_m+1$.
        \Else 
        \State Select a random $R'\in[0.1,0.3)$, $C_s=C_s+1$.
        \EndIf
        \ElsIf{$R\in[0.3,0.7)$}
        \If{$N_m<C_m$} 
        \State Select a random $R'\in[0.3,0.7)$, $C_m=C_m+1$.
        \Else \State Select a random $R'\in[0.1,0.3)$, $C_s=C_s+1$.
        \EndIf
        \Else
        \State Select a random $R'\in[0.1,0.3)$, $C_s=C_s+1$.
        \EndIf
        \State Resize $(I_i^{(ca)}, y_i)$ to $(I_i^{(ca)}{}', y'_i)$ by the ratio $r=R'/R$.
        \State Get the resized anomaly pattern $A'_i=I_i^{(ca)}{}'(y'_i)$.
        \State Select the injection location $l_c\in \mathcal{S}$ by Eq. (\ref{eq:inloc}).
        \State Get the pseudo anomaly $I^{(sa)}=PEI(I^{(n)}, y^{(n)}, l_c, A'_i)$.
        \State Compute $y^{(sa)}$ by Eq. (\ref{eq:mask}), $\mathcal{I}^{(sa)}=\mathcal{I}^{(sa)}\cup (I^{(sa)}, y^{(sa)})$.
        \EndWhile
        \State \textbf{return} $\mathcal{I}^{(sa)}$
    \end{algorithmic}
\end{algorithm}

\textbf{Multi-scale Anomaly Synthesis (MAS).}
To promote the diversity in pseudo anomalies' scale, CAI adopts a Multi-scale Anomaly Synthesis strategy. Given the target image, the number of large/medium/small pseudo anomalies, $N_l/N_m/N_s$, are explicitly set for anomaly synthesis. Meanwhile, MAS requires prioritizing synthesis of larger pseudo anomalies under the constraints of STM. For example, MAS prioritizes synthesizing large pseudo anomalies when given a large anomaly pattern, while medium or small pseudo anomalies are produced only when the required number of large pseudo anomaly is reached. Therefore, an anomaly pattern prefers synthesizing a pseudo anomaly of the same scale, but it can also synthesize pseudo anomalies in other scales. In this way, a rational $R'$ is selected by STM and MAS (see line 6-23 of Algo. \ref{alg:cai}). In this paper, we synthesize $N_l=4, N_m=3, N_s=3$ pseudo anomalies with a target normal image, which is a relatively balanced ratio for performance and computational overhead. 

\textbf{Injection Location Selection (ILS).} The injection location $l=(h,w)$ is defined as the center of injected anomaly pattern $A'_i$ in the target image $I^{(n)}$, where $h, w$ are coordinates of the target image $I^{(n)}$. We require a legal $l$ to satisfy two conditions: (1) Since anomalies on the foreground are more meaningful, $l$ should fall into $I^{(n)}$'s foreground region formed by non-zero pixels in $y^{(n)}$. (2) $l$ should allow a complete injection of the resized anomaly pattern $A'_i$. Specifically, when putting the center of $A'_i$'s minimum bounding box on this location, the anomaly pattern $A_i$ will not exceed the range of $I^{(n)}$ (see Fig. \ref{fig:CAI}). By two conditions above, we determine a candidate set $\mathcal{S}$ of injection location as follows:

\begin{equation}
\label{eq:inloc}
\begin{aligned}
    & \mathcal{S}_1 = \{l|y^{(n)}(l)=1\} \\
    & \mathcal{S}_2 = \{l|CI(I^{(n)}, \Omega(I^{(n)}, l, A'_i))=True\} \\
    & \mathcal{S}=\mathcal{S}_1\cap\mathcal{S}_2
\end{aligned}
\end{equation}
where $\Omega(I^{(n)}, l, A'_i)$ refers to the region that the injected anomaly pattern $A'_i$ occupies when it is centered at $(h,w)$ in $I^{(n)}$, and the function $CI(\cdot)$ judges whether $\Omega(I^{(n)}, l, A'_i)$ is completely included by $I^{(n)}$. Eventually, we randomly select a location $l_c\in \mathcal{S}$ to obtain diverse anomaly locations. 

\textbf{Poisson Editing (PE) based Anomaly Injection.} 
With the resized anomaly pattern $A'_i$ and selected injection location $l_c$, we can perform anomaly injection. Most existing model-free ZSAS solutions perform injection by direct pasting \cite{li2021cutpaste, zavrtanik2021draem} or smooth blending \cite{zou2022spot}, which cause salient artifacts (e.g. pixel discontinuity or blur) and degrade pseudo anomalies' authenticity. Moreover, artifacts provide a ``shortcut'' for training and distract the IAD model from learning useful features of anomalies. Therefore, we are inspired by PE \cite{10.1145/1201775.882269,schluter2022natural} and introduce to seamlessly inject anomaly patterns into target images. Concretely, PE formulates anomaly injection into the following problem:

\begin{equation}
\label{eq:poission_edit_1}
\begin{aligned}
& \quad\min_{I^{(sa)}}\int\int_{\Omega(I^{(n)}, l_c, A'_i)}|\nabla I^{(sa)}-\nabla A'_i|^2, \\
& \quad s.t.\quad I^{(sa)}|_{\partial\Omega(I^{(n)}, l_c, A'_i)}=I^{(n)}|_{\partial\Omega(I^{(n)}, l_c, A'_i)}  
\end{aligned}
\end{equation}
where $\partial\Omega(I^{(n)}, l_c, A'_i)$ denotes the border of region $\Omega(I^{(n)}, l_c, A'_i)$. Eq. (\ref{eq:poission_edit_1}) actually encourages $\Omega(I^{(n)}, l_c, A'_i)$ to share the pattern of the injected anomaly $A'_i$ by minimizing their difference in gradients, while the constraint of Eq. (\ref{eq:poission_edit_1}) ensures a smooth transition at the region border. Eq. (\ref{eq:poission_edit_1}) is equivalent to solving a Poisson partial differential equation that subjects to Dirichlet border condition.  Two modes can be selected in PE: For \textit{Normal} mode, the injected pattern is solely fetched from the source image; For \textit{Mixed} mode, the injected pattern is a mixture of source and target image pattern. We denote CAI using \textit{Normal} and \textit{Mixed} mode PE by CAI-N and CAI-M, respectively. Then, we filter out the injected anomaly pattern outside the foreground region by the foreground mask $y^{(n)}$. Considering all inputs, the whole PE-based injection process can be denoted by $I^{(sa)}=PEI(I^{(n)}, y^{(n)}, l_c, A'_i)$. 


Finally, to obtain the anomaly mask $y^{(sa)}$ for each pseudo anomaly image $I^{(sa)}$, we follow \cite{schluter2022natural} to compute a pixel-wise difference map $I^{(d)}=|I^{(sa)}-I^{(n)}|$, and $y^{(sa)}$ can be obtained by binarizing $I^{(d)}$ with a threshold $T$:

\begin{equation}
\label{eq:mask}
    y^{(sa)}=bin(I^{(d)}, T)
\end{equation}

\begin{figure}[t]
  \centering
   \includegraphics[width=1.\linewidth]{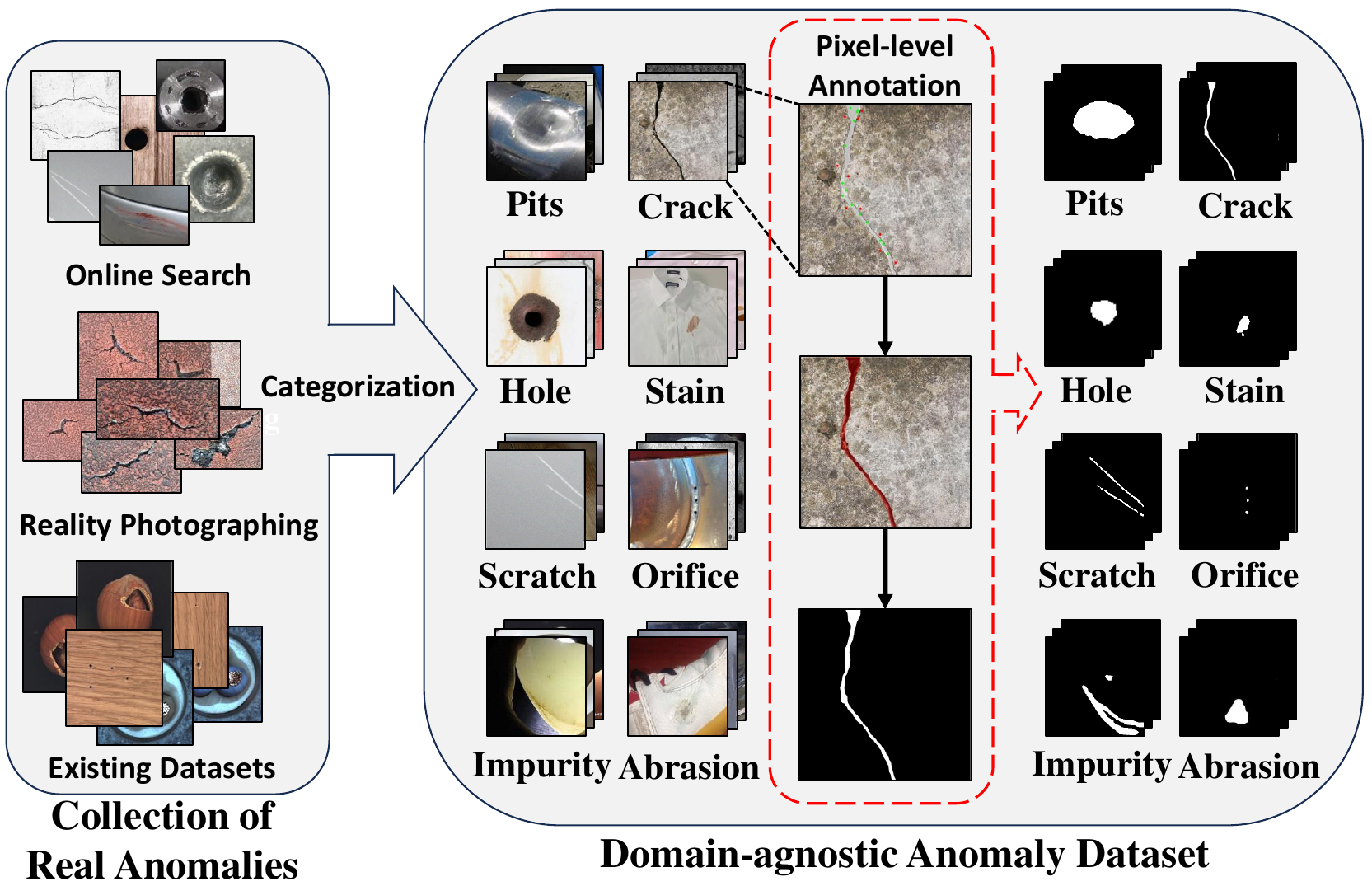} 
   \caption{Procedure to build our domain-agnostic anomaly dataset.}
   \label{fig:oad}
\end{figure}

\subsection{Domain-agnostic Anomaly Dataset (DAAD)}

One of CAI's foundation is a real anomaly dataset that can provide abundant cross-domain anomalies. However, we notice that existing IAD datasets typically follow a domain-oriented fashion, i.e. they focus on collecting normal/abnormal images under certain domains, and anomalies are mainly categorized by their associated domain. For example, for the popular MvTec-AD dataset \cite{bergmann2019mvtec}, its primary categories are like ``\textit{Bottle}/\textit{Capsule}/\textit{Tile}...''. Unfortunately, this fashion suffers from two major drawbacks: (1) It hinders the collection of extensive real anomalies, as only anomalies in a certain domain are of interest. (2) It stresses the domain rather than anomalies. Nevertheless, CAI requires massive cross-domain anomalies, which makes existing IAD datasets short in quantity and diversity of real anomalies. This inspires us to build the first domain-agnostic anomaly dataset named DAAD, and we show its building procedure in Fig. \ref{fig:oad} and detail its key steps below.

\textbf{Real Anomaly Collection.} 
Real anomalies are actually abundant when the constraint of domain is removed. To collect domain-agnostic real anomalies,  we exploit three sources that are all simple and affordable for the general public: (1) \textit{Online search}: We feed IAD-related textual queries (e.g. ``industry scratch'') into the search engine to obtain real anomaly images. (2) \textit{Reality photographing}: We use frequently-seen portable photographing devices like digital cameras or cellphones to capture real anomalies. (3) \textit{Existing datasets}: We exploit real anomalies from 9 existing IAD datasets (listed in supplementary material). By virtue of three sources above, we are able to collect about 6000 real anomaly images in a short period ($\sim$ 2 weeks) as the data foundation of our domain-agnostic anomaly dataset. Our anomaly collection is much more extensible when compared with classic IAD datasets, as DAAD welcomes any real anomaly from diverse domains. Note that if an existing dataset is used for evaluation, its data will be strictly excluded from ZSAS to prevent information leakage.

\textbf{Anomaly Categorization and Annotation.} After collecting sufficient real anomalies, we categorize real cross-domain anomalies into several anomaly classes $\{\mathcal{C}_1, \mathcal{C}_2,...\}$ in an anomaly-oriented manner, which is a different practice from mainstream IAD datasets. Anomaly categorization benefits ZSAS/IAD in terms of two aspects: (1) It enables us to appoint a certain anomaly class for ZSAS, which can be vital in some domains where certain classes of anomalies occur by a higher frequency. (2) It facilitates us to learn characteristics of a certain anomaly class by grouping them together. Here we categorize collected anomalies into 8 frequently-encountered industrial anomaly classes: \textit{Pits}, \textit{Crack}, \textit{Hole}, \textit{Stain}, \textit{Scratch}, \textit{Orifice}, \textit{Impurity}, \textit{Abrasion}. Having categorized collected anomalies, we ensure each anomaly image is assigned with a mask to indicate its anomaly pattern. For images from online search, reality photographing and datasets without masks, we annotate them by an annotation tool named EISeg \cite{hao2022eiseg}, while those anomaly masks from existing datasets can be directly fetched. In this way, our domain-agnostic anomaly dataset can be built in an efficient and extendible manner. DAAD is open and its details are shown in supplementary material.

\subsection{CAI-guided Diffusion Mechanism (CDM)}


As shown above, our model-free CAI enables both efficient and effective ZSAS, but it relies on DAAD dataset. Although our DAAD dataset has already provided considerable real anomaly patterns, its further extension requires time and labor, which can be an impediment to CAI-based ZSAS. To break the quantity limit of real anomalies in DAAD, a natural alternative is to exploit high-quality pseudo anomalies yielded by CAI to train a generative model, so it can enable unlimited ZSAS and generate more diverse pseudo anomalies. To this end, we propose CAI-guided Diffusion Mechanism (CDM), which serve as a powerful model-based supplement alongside our model-free CAI solution. CDM is inspired by Latent Diffusion Model (LDM) \cite{rombach2022high}, Textual Inversion (TI) \cite{DBLP:conf/iclr/GalAAPBCC23} and a recent FSAS solution \cite{hu2024anomalydiffusion}. It is trained by the objective below:
\begin{equation}
    \begin{aligned}
      e^*, E^*_s = & \arg\min_{e, E_s}\mathbb{E}_{z_i\sim E_a(I_i^{(sa)}), y_i^{(loc)},\epsilon, t}\mathcal{L}_{TI}, \\
      \mathcal{L}_{TI} = & ||y_i^{(loc)}\odot (\epsilon - \epsilon_\theta(z_i^t, t, e))||^2_2
    \end{aligned}
\end{equation}
Due to page limit, see supplementary material for CDM's details. Guided by CAI's high-quality pseudo anomalies, CDM can generate more diverse pseudo anomalies with good authenticity (see Fig. \ref{fig:CAI-guided Diffusion}), while experiments in Sec. \ref{sec:results} also show that pseudo anomalies generated by CDM can be readily applied to training a IAD model. Note that CAI is independent of CDM, as CDM is a supplement to CAI.

    
\section{Experiment}

\subsection{Experimental Setup}

\textbf{Evaluation Protocol.} To conduct a dedicated evaluation on ZSAS, our evaluation protocol is designed based on \cite{hu2024anomalydiffusion}, the latest work that has conducted a specialized evaluation of anomaly synthesis. Concretely, pseudo anomalies are evaluated in terms of two facets: (1) \textit{Utility in IAD}. Since the goal of ZSAS is to serve IAD here, we evaluate the utility of different ZSAS solutions' pseudo anomalies by a unified IAD framework: With each target image from a normal training set, ten pseudo anomaly images are synthesized. Then, a binary-class training set is built by combining given normal images and pseudo anomalies. Since our goal is to manifest the performance difference of various ZSAS solutions rather than pursuing SOTA performance on a certain dataset, we only train a basic UNet from \cite{zavrtanik2021draem,hu2024anomalydiffusion} by Focal loss \cite{ross2017focal}, and avoid introducing more modules or boosting tricks. Following \cite{hu2024anomalydiffusion}, with the pixel-level anomaly mask output by the trained UNet, IAD performance is evaluated by three metrics: Pixel-level Area under Receiver Operating Characteristic Curve (AUC), Average Precision (AP) and Per-Region Overlap (PRO). For each ZSAS solution, we select the best-performing model among 200 training epochs for comparison to avoid setting optimal epoch number. The training process of all ZSAS solutions shares the same hyperparameters and random seed (see supplementary material). (2) \textit{Authenticity}. We measure authenticity by FID \cite{heusel2017gans}: Lower FID reflects better similarity between pseudo anomalies and domain-specific anomalies of the test set.  

\textbf{Benchmark Datasets and Competing Methods.} We use three widely-used IAD benchmark datasets: MvTec-AD \cite{bergmann2019mvtec}, VisA \cite{zou2022spot} and KSDD2 \cite{bovzivc2021mixed}. Since MvTec-AD and VisA are included into our domain-agnostic anomaly dataset, it should be noted that \textit{\textbf{they will be excluded from ZSAS when experiments are conducted on themselves to avoid information leakage}}. As to competing ZSAS methods, we compare the proposed CAI with 5 SOTA ZSAS solutions: CutPaste \cite{li2021cutpaste}, NSA \cite{schluter2022natural}, DRAEM \cite{zavrtanik2021draem}, GLASS \cite{chen2024unified} and RealNet \cite{zhang2024realnet}. Since NSA also employs Poisson editing, it is also categorized into NSA-N and NSA-M, like our CAI-N and CAI-M. Besides, we also include the latest FSAS solution AnomalyDiffusion (ADiff) \cite{hu2024anomalydiffusion} as a reference. More details of experimental setup are given in supplementary material.

\subsection{Experimental Results}
\label{sec:results}

\begin{figure}[t]
  \centering
   \includegraphics[width=1\linewidth]{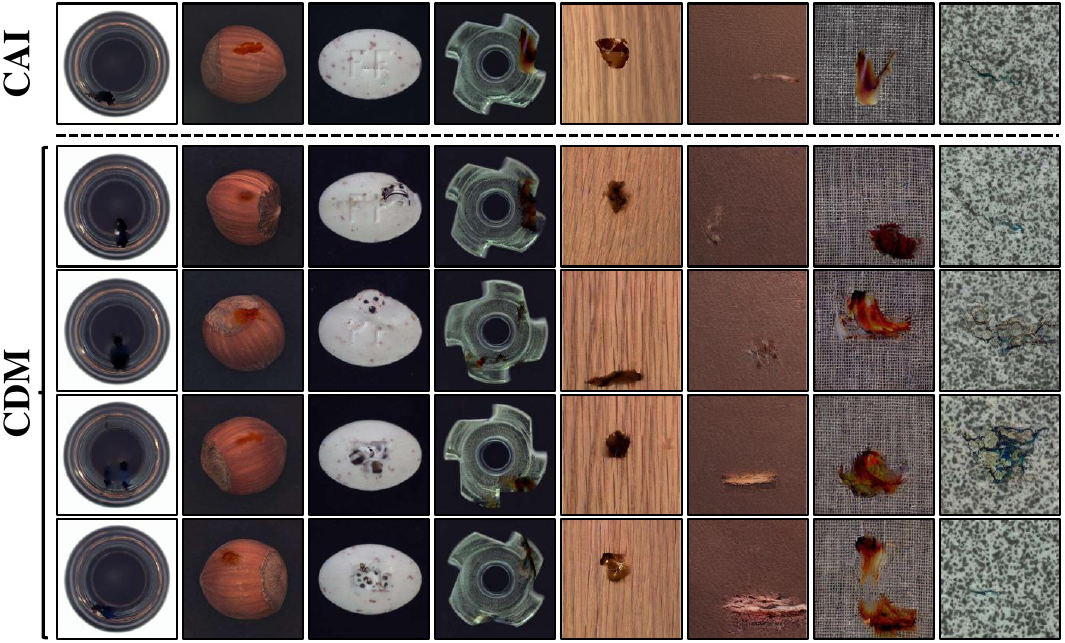} 
   \caption{Pseudo anomalies of CAI (row 1) and CDM (row 2-5).}
   \label{fig:CAI-guided Diffusion}
\end{figure}

\begin{table}[t]
\caption{Pixel-level IAD performance ($\%$) of existing ZSAS solutions. The best/2nd best performer is in boldface/underlined. }
\centering
\resizebox{\columnwidth}{!}{%
\begin{tabular}{l|ccc|ccc|ccc}
\toprule[1.2pt]
 & \multicolumn{3}{c|}{MvTec-AD} & \multicolumn{3}{c|}{VisA} & \multicolumn{3}{c}{KSDD2} \\ \cmidrule{2-10}
 & AUC & AP & PRO & AUC & AP & PRO & AUC & AP & PRO \\ \midrule
CutPaste & 89.6  & 38.5  & 71.9  & 93.0  & 15.8  & 72.2  & 89.0  & 39.6  & 70.9   \\
NSA-N & 92.4 & 49.1 & 80.6 & 95.7 & 37.3 & 83.9 & 80.5 & 4.5  & 55.5 \\
NSA-M & 91.2  & 47.2  & 78.8  & 94.5  & 25.9  & 83.4  & 87.6  & 34.8  & 69.9   \\
DRAEM & 91.0  & 54.5  & 82.5  & 93.2  & 37.3  & 85.6  & 93.8  & 52.9  & \ul{86.6}   \\
RealNet & 85.1  & 39.0  & 70.9  & 85.5  & 21.8  & 79.1  & 81.3  & 18.9  & 64.3   \\
GLASS & 91.1  & 52.0  & 82.3  & 91.4  & 27.0  & 79.1  & 92.5  & 42.7  & 82.3   \\
\midrule
\midrule
\cellcolor[HTML]{FFFFEC} CAI-N & \cellcolor[HTML]{FFFFEC} \textbf{95.6} & \cellcolor[HTML]{FFFFEC} \textbf{62.6} & \cellcolor[HTML]{FFFFEC} \textbf{87.5} & \cellcolor[HTML]{FFFFEC} \textbf{96.1} & \cellcolor[HTML]{FFFFEC} \ul{39.7} & 
 \cellcolor[HTML]{FFFFEC} \ul{87.7} & \cellcolor[HTML]{FFFFEC} \ul{96.6} & \cellcolor[HTML]{FFFFEC} \textbf{70.4} & \cellcolor[HTML]{FFFFEC} \textbf{89.6} \\ 
 
 \cellcolor[HTML]{FFFFEC} CAI-M & \cellcolor[HTML]{FFFFEC} \ul{94.7} & \cellcolor[HTML]{FFFFEC} \ul{61.6} & \cellcolor[HTML]{FFFFEC} \ul{86.6} & \cellcolor[HTML]{FFFFEC} \ul{95.9} & \cellcolor[HTML]{FFFFEC} \textbf{40.1} & 
 \cellcolor[HTML]{FFFFEC} \textbf{89.1} & \cellcolor[HTML]{FFFFEC} \textbf{96.8} & \cellcolor[HTML]{FFFFEC} \ul{66.3} & \cellcolor[HTML]{FFFFEC} \textbf{89.6} \\ \bottomrule[1.2pt]
\end{tabular}%
}
\label{tab:iad_results}
\end{table}

\textbf{Quality of Pseudo Anomalies.} We first compare the quality of pseudo anomalies from different ZSAS solutions in Table \ref{tab:iad_results}-\ref{tab:fid}. For utility in IAD, pseudo anomalies synthesized by CAI constantly achieve the best IAD performance in terms of all metrics on all datasets (see Table \ref{tab:iad_results}). Meanwhile, CAI-N (\textit{Normal} mode PE) achieves better overall performance than CAI-M  (\textit{Mixed} mode PE). Compared with SOTA ZSAS counterparts, CAI brings notable improvement in almost all cases, while it even outperforms the 2nd best performer by 17.5\% AP on KSDD2. Such results verify the effectiveness of CAI for ZSAS in IAD. As to authenticity, pseudo anomalies synthesized by CAI-M and CAI-N attain the lowest and 2nd lowest FID on all 3 datasets (see Table \ref{tab:fid}), even though NSA and CutPaste actually enjoy an unfair advantage in FID calculation: They sample normal image patches of the train set as anomaly patterns, which are naturally similar to the test set of the same domain. By contrast, without using patches of the same domain as anomalies, CAI's FID is superior to NSA and CutPaste, which reveals that CAI can synthesize highly authentic pseudo anomalies for the specific domain. In addition, we also have two interesting observations: First, CutPaste's FID is higher than NSA, which can be ascribed to image artifacts caused by pasting-based anomaly injection. Second, although model-based RealNet yields lower FID than model-free methods DRAEM/GLASS on MvTec-AD/VisA dataset, it performs poorly on KSDD2 dataset, whose images have evidently different characteristics from MvTec-AD/VisA. Such results also reveal the complexity in designing the tricky generative process for various domains. As a summary, CAI can synthesize high-quality pseudo anomalies in a simple and training-free manner.  

\begin{figure}[t]
  \centering
   \includegraphics[width=1\linewidth]{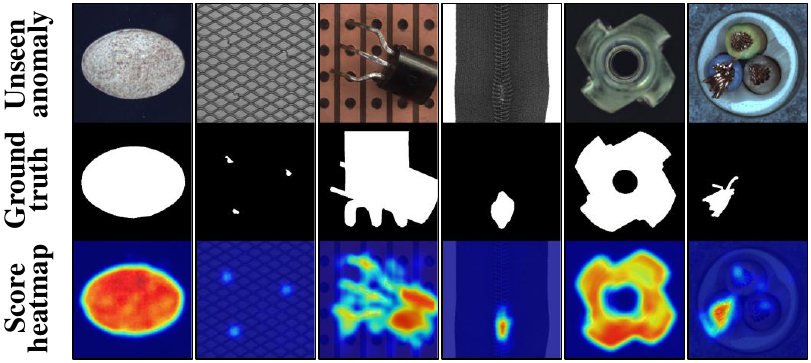} 
   \caption{IAD detector trained by pseudo anomalies synthesized by CAI and DAAD can generalize to different unseen anomalies, e.g. wrong pill type, bent grid, transistor misplacement.} 
   \label{fig:Generalized demonstration}
\end{figure}

\begin{table}[t]
\caption{FID of pseudo anomalies for different ZSAS solutions.}
\centering
\resizebox{\columnwidth}{!}{%
\begin{tabular}{l|ccccccccc}
\toprule[1.2pt]
 & CutPaste  & NSA-N & NSA-M & DRAEM & RealNet & GLASS & \cellcolor[HTML]{FFFFEC} CAI-N & \cellcolor[HTML]{FFFFEC} CAI-M \\
\midrule
MvTec-AD & 153.4 & 138.4 & 136.9 & 182.6  & 139.5  & 175.6  & \cellcolor[HTML]{FFFFEC} \ul{121.8}  & \cellcolor[HTML]{FFFFEC} \textbf{107.0}   \\ 
VisA & 98.5 & 78.1 & 74.9 & 99.3  & 81.7  & 93.8  & \cellcolor[HTML]{FFFFEC} \ul{45.9}  & \cellcolor[HTML]{FFFFEC} \textbf{37.9}   \\ 
KSDD2 & 201.9 & 187.0 & 185.2 & 244.9  & 301.2  & 224.9  & \cellcolor[HTML]{FFFFEC} \ul{169.4}  & \cellcolor[HTML]{FFFFEC} \textbf{147.9}   \\ \bottomrule[1.2pt]
\end{tabular}%
}
\label{tab:fid}
\end{table}

\textbf{Pseudo Anomaly Extension.} Next, we validate CDM for pseudo anomaly extension by results in Fig. \ref{fig:CAI-guided Diffusion} and Fig. \ref{fig:cai_cdm}. Due to page limit, we report results on MvTec-AD in subsequent discussion to unveil the trend. We first compare CDM's generated anomalies and CAI's synthesized anomalies: As shown in Fig. \ref{fig:CAI-guided Diffusion}, pseudo anomalies produced by both CAI and CDM enjoy satisfactory authenticity. Meanwhile, CDM can generate pseudo anomalies with similar anomaly appearance and diverse locations. For further justification, we directly use CDM's extended pseudo anomalies as training data of the UNet detector and evaluate its performance. We test two configurations:  Training with pseudo anomalies from all 8 anomaly classes (CDM-8) or 5 anomaly classes that are more likely to occur in the target domain (CDM-5). As shown in Fig. \ref{fig:cai_cdm}, the detector trained by extended pseudo anomalies achieves surprisingly close IAD performance to CAI's synthesized pseudo anomalies. Interestingly, CDM-5 achieves better performance than CDM-8, while it even surpasses CAI in terms of AUC. Such results justify anomaly categorization, which allows us to appoint the anomaly class for ZSAS. Thus, CDM can serve as a powerful way to extend pseudo anomalies. 

\textbf{Generalization to Unseen Anomalies.} As discussed above, the anomaly categories in DAAD dataset are still finite. Thus, one may wonder whether pseudo anomalies synthesized by CAI and DAAD dataset can generalize to unseen anomaly categories. For example, on MvTec-AD dataset, bent wire and accessory misplacement are not included in our DAAD dataset. However, as shown by Fig. \ref{fig:Generalized demonstration}, the IAD detector, which is trained by pseudo anomalies synthesized by our CAI and DAAD datasets, can effectively detect those unseen anomalies. This demonstrates that CAI enables the detector to generalize to unseen anomalies. 



\begin{figure}[t]
    \centering
    \begin{subfigure}[b]{0.22\textwidth}
        \includegraphics[width=\textwidth]{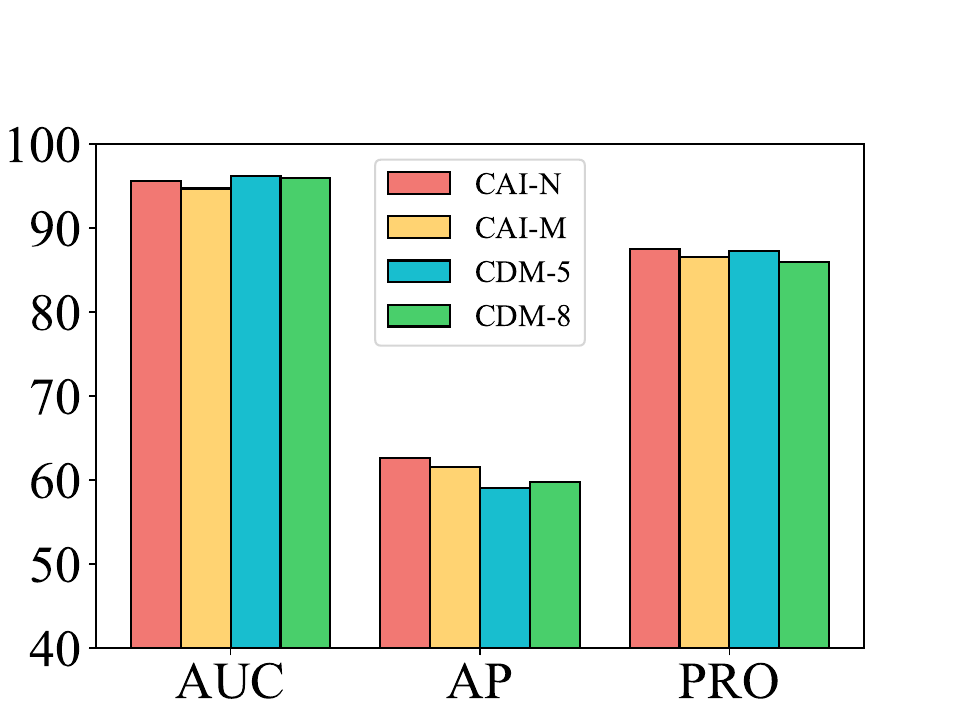}
        \caption{CAI vs. CDM.}
        \label{fig:cai_cdm}
    \end{subfigure}
    \begin{subfigure}[b]{0.22\textwidth}
        \includegraphics[width=\textwidth]{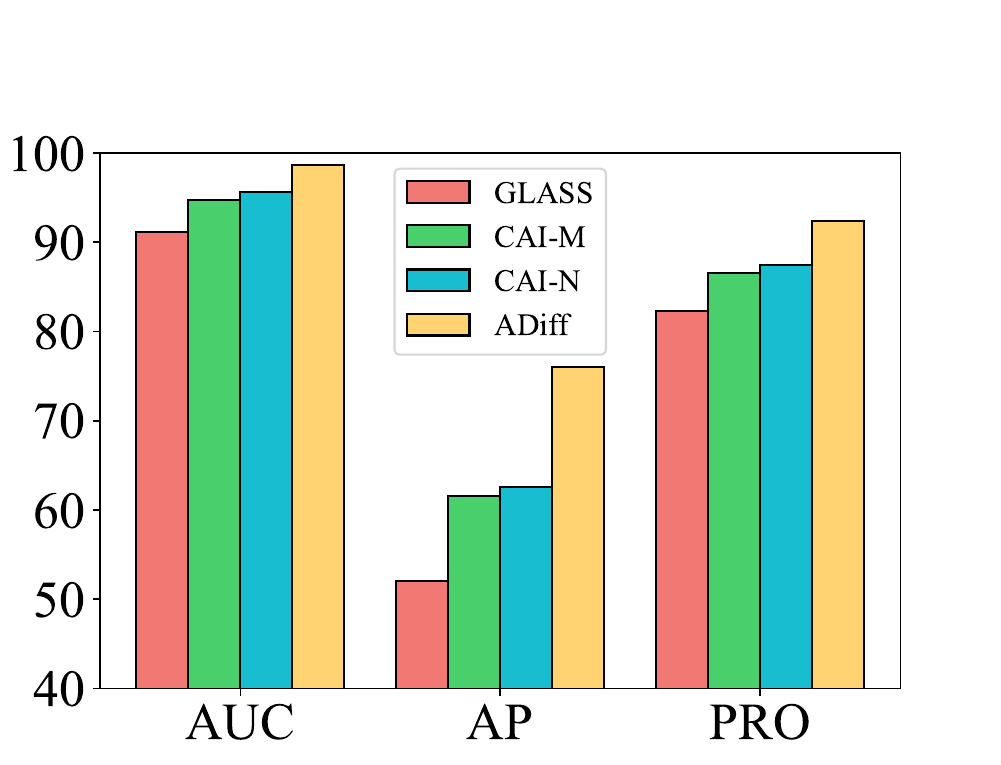}
        \caption{FSAS vs. ZSAS.}
        \label{fig:fsas_zsas}
    \end{subfigure}

    \begin{subfigure}[b]{0.22\textwidth}
        \includegraphics[width=\textwidth]{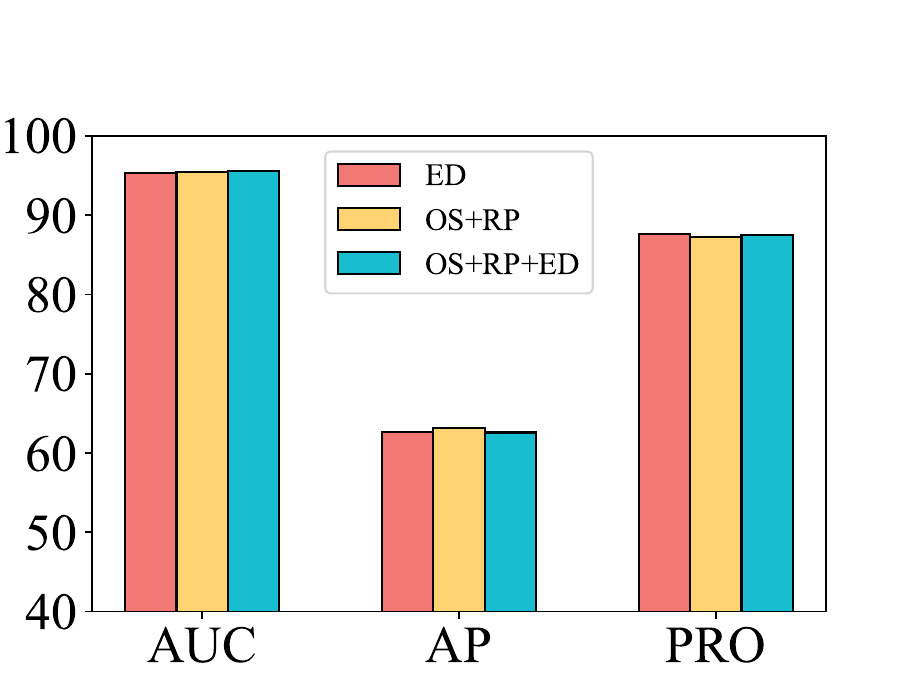}
        \caption{Different anomaly sources.}
        \label{fig:source}
    \end{subfigure}
    \begin{subfigure}[b]{0.22\textwidth}
        \includegraphics[width=\textwidth]{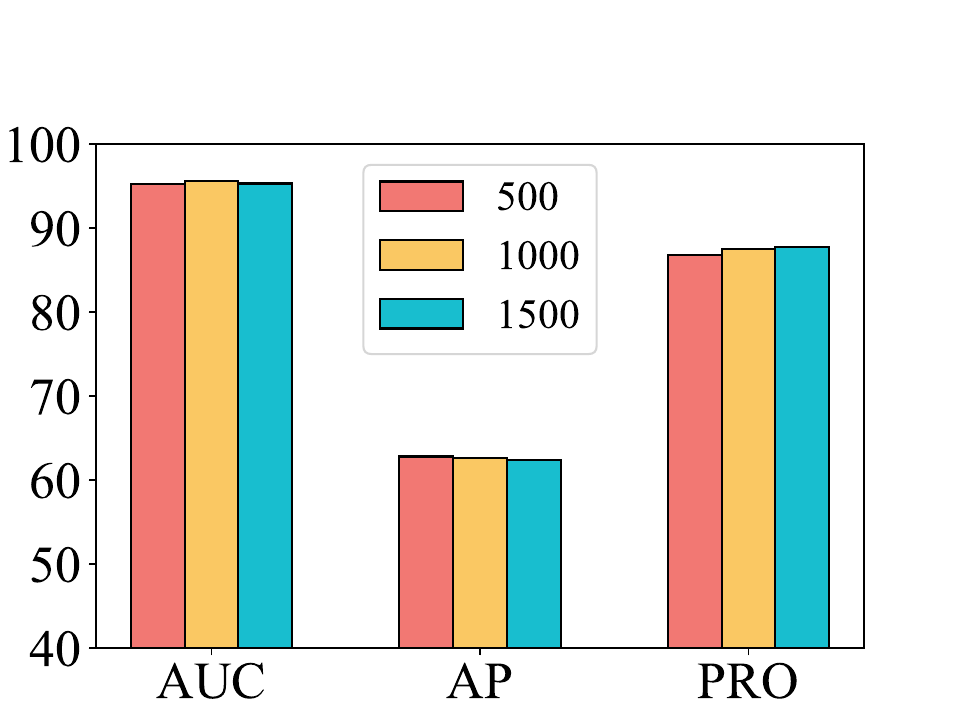}
        \caption{Different sampling number.}
        \label{fig:unet_sample}
    \end{subfigure}
    \caption{Performance (\%) comparison of different configurations.}
    \label{fig:imageset}
\end{figure}

\textbf{Ablation Studies.} In Table \ref{tab:ablation}, we conduct ablation studies to verify the role of each technique in CAI-N: (1) Without PE-based injection (line 1), the performance is degraded by over $5\%$ under all metrics, which unveils its importance in eliminating artifacts. (2) The removal of Source-target matching (STM) leads to tangible performance loss in terms of all three metrics (line 2), which can be up to 0.8\%. This verifies the necessity of STM. (3) Injection location selection (ILS) provides a more precise and valid injection position, enhancing performance by up to 1.6\% (line 3). 


\textbf{Discussion.} (1) We first compare the cases where different ratios of large, medium and small pseudo anomalies are used for CAI: Large-biased (6:2:2), medium-biased (2:6:2), small-biased (2:2:6) and balanced (4:3:3). As shown in Table \ref{tab:ablation_msa}, the balanced case achieves the better overall performance than any of the biased cases. This manifests the importance to perform a muti-scale anomaly synthesis that considers all scales fairly.
(2) \textit{FSAS vs. ZSAS}: We also compare ZSAS solutions (CAI and GLASS \cite{chen2024unified}) with the recent FSAS solution ADiff \cite{hu2024anomalydiffusion}. ADiff is a strong FSAS competitor, as it exploits domain-specific anomalies from all anomaly classes of the test set as training data. As shown in Fig. \ref{fig:fsas_zsas}, although the performance gap still exists, CAI has evidently narrowed the gap between ZSAS and FSAS when compared with the latest GLASS. (3) \textit{Data sources}. We collect cross-domain anomalies from three sources: Online search (OS), reality photographing (RP) and existing datasets (ED), while images from OS and RP are newly collected. We compare three data source configurations: OS+RP+ED, OS+RP and ED only. As shown in Fig. \ref{fig:source}, three configurations actually achieve comparable performance, which suggests that real anomalies from different sources can be readily transferred to a specific domain, while this also verifies the quality of our newly-collected anomaly data. (4) \textit{Number of UNet detector sampling.} We adjusted the number of samples used in training the UNet detector, reporting scenarios with 500 and 1500 samples in CAI. As shown in Fig. \ref{fig:unet_sample}, it was observed that the model's performance does not undergo drastic change, so we simply choose 1000 by default. (5) \textit{Computational cost.} It takes an average 0.019s and 0.018s for CAI-N and CAI-M to synthesize one pseudo anomaly image on the MvTec-AD dataset, which are fairly acceptable. Due to page limit, more results on cost are given in supplementary material.

\begin{table}[t]
\small 
\caption{Results of ablation studies on core components of CAI.}
\centering
\resizebox{0.6\columnwidth}{!}{
\begin{tabular}{ccc|ccc}
\toprule[1.2pt]
PE & STM & ILS & AUC & AP & PRO\\
\midrule
- & \checkmark & \checkmark & 89.7  & 50.1  & 77.5 \\
\checkmark & - & \checkmark & 95.0  & 61.8  & 86.7 \\
\checkmark & \checkmark & - & 95.1  & 61.3  & 85.9 \\
\checkmark & \checkmark & \checkmark & \textbf{95.6}  & \textbf{62.6} & \textbf{87.5} \\
\bottomrule[1.2pt]
\end{tabular}
}
\label{tab:ablation}
\end{table}

\begin{table}[t]
\small 
\caption{Discussion on multi-scale anomaly synthesis.}
\centering
\resizebox{0.62\columnwidth}{!}{
\begin{tabular}{l|lll}
\toprule[1.2pt]
 & AUC & AP & PRO\\
\midrule
Large-biased (6:2:2) & 95.2 & 61.7 & 86.8 \\
Medium-biased (2:6:2) & 94.8 & 61.5 & 86.7 \\
Small-biased (2:2:6) & 95.4 & 61.8 & \textbf{87.8} \\
Balanced (4:4:3) & \textbf{95.6}  & \textbf{62.6}  & 87.5 \\
\bottomrule[1.2pt]
\end{tabular}
}
\label{tab:ablation_msa}
\end{table}

\section{Conclusion}

In this paper, we propose a brand-new paradigm that can ``use stones from other mountains to polish jade'', i.e. exploiting abundant cross-domain anomalies to solve the ZSAS problem under a specific domain. Our paradigm is supported by a novel ZSAS method named CAI, a newly-constructed domain-agnostic anomaly dataset and a CAI-guided diffusion mechanism for flexible anomaly extension. Based on our paradigm, we can realize highly-effective ZSAS in a training-free and extendible manner, while the synthesized pseudo anomalies are able to achieve superior IAD performance and higher authenticity. The major limitation is that our paradigm focuses on appearance anomalies in this paper, while we will also explore the synthesis of logical anomalies \cite{zhang2024contextual} in the future.
{
    \small
    \bibliographystyle{ieeenat_fullname}
    \bibliography{main}
}



\end{document}